\documentclass[12pt]{article}

\usepackage{url}
\usepackage{amsfonts}
\usepackage{amsmath,amssymb,color}
\usepackage{graphicx}
\usepackage{enumitem}
\usepackage{multirow}
\usepackage{bm}
\usepackage{natbib}
\usepackage{verbatim}
\usepackage{float}
\usepackage[toc,page]{appendix}
\usepackage{setspace,fullpage}
\usepackage{setspace}
\usepackage{adjustbox}
\usepackage{rotating}
\usepackage{subcaption}
\usepackage{amsthm}
\usepackage{booktabs,dcolumn,caption}
\usepackage{accents}
\usepackage{algpseudocode}
\usepackage{algorithm}

\usepackage{tikz}
\usetikzlibrary{shapes.geometric, arrows}

\tikzstyle{startstop} = [rectangle, rounded corners, 
minimum width=1.5cm, 
minimum height=1cm,
text centered, 
draw=black]

\tikzstyle{io} = [trapezium, 
trapezium stretches=true, 
trapezium left angle=70, 
trapezium right angle=110, 
minimum width=3cm, 
minimum height=1cm, text centered, 
draw=black, fill=blue!30]

\tikzstyle{process} = [rectangle, 
minimum width=3cm, 
minimum height=1cm, 
text centered, 
text width=3cm, 
draw=black, 
fill=orange!30]

\tikzstyle{decision} = [diamond, 
minimum width=2cm, 
minimum height=1cm, 
text centered, 
draw=black]
\tikzstyle{arrow} = [thick,->,>=stealth]

\usepackage{xr}
\usepackage{array}
\newcolumntype{L}[1]{>{\raggedright\let\newline\\\arraybackslash\hspace{0pt}}p{#1}}
\newcolumntype{C}[1]{>{\centering\let\newline\\\arraybackslash\hspace{0pt}}p{#1}}
\newcolumntype{R}[1]{>{\raggedleft\let\newline\\\arraybackslash\hspace{0pt}}p{#1}}

\newcommand{\yc}[1]{{{\color{red}  #1}}}

\newcommand{\ddd}{\boldsymbol \delta}

\newcommand{\ttt}{\boldsymbol \theta}
\newcommand{\TTT}{\boldsymbol \Theta} 

\newcommand{\mm}{\boldsymbol \mu}
\newcommand{\hhh}{\boldsymbol \eta}
\newcommand{\tttt}{\boldsymbol \tau}

\newcommand{\CCC}{\mathcal C}

\newcommand{\DDD}{\Delta}

\newcommand{\EEE}{\mathbb E}

\newcommand{\JJJ}{\mathcal J}

\newcommand{\MM}{\mathbf M}
\newcommand{\MMM}{\mathcal M}

\newcommand{\YYY}{\mathcal Y}

\newcommand{\II}{\mathbf I}

\newcommand{\LLL}{\mathcal L}

\newcommand{\OOO}{\mathbb O}

\newcommand{\PPP}{\mathcal P}
\newcommand{\R}{\mathbb R}

\newcommand{\SSSS}{\mathcal S}
\newcommand{\TT}{\mathcal T}

\newcommand{\uu}{\mathbf u}
\newcommand{\UU}{\mathbf U}
\newcommand{\VV}{\mathbf V}
\newcommand{\vv}{\mathbf v}
\newcommand{\WW}{\mathbf W}
\newcommand{\ww}{\mathbf w}

\newcommand{\1}{\uppercase\expandafter{\romannumeral1}}
\newcommand{\2}{\uppercase\expandafter{\romannumeral2}}

\newcommand{\rank}{\text{rank}}

\newcommand{\opt}{\text{opt}}
\newcommand{\str}{\text{str}}

\newcommand{\argmin}{\operatornamewithlimits{arg\,min}}

\newtheorem{theorem}{Theorem}

\newtheorem{corollary}{Corollary}

\newtheorem{proposition}{Proposition}
\newtheorem{remark}{Remark}

\title{Efficient Estimation for Longitudinal Networks via Adaptive Merging}
\author{Haoran Zhang$^\dag$ and Junhui Wang$^\ddag$\\ [10pt]
	$^\dag$ Department of Statistics and Data Science \\
	Southern University of Science and Technology 
	\and
	$^\ddag$Department of Statistics \\
	The Chinese University of Hong Kong
}
\date{}

\begin{document}
\maketitle

\onehalfspacing


\begin{abstract}
Longitudinal network consists of a sequence of temporal edges among multiple nodes, where the temporal edges are observed in real time. It has become ubiquitous with the rise of online social platform and e-commerce, but largely under-investigated in literature. In this paper, we propose an efficient estimation framework for longitudinal network, leveraging strengths of adaptive network merging, tensor decomposition and point process. It merges neighboring sparse networks so as to enlarge the number of observed edges and reduce estimation variance, whereas the estimation bias introduced by network merging is controlled by exploiting local temporal structures for adaptive network neighborhood. A projected gradient descent algorithm is proposed to facilitate estimation, where the upper bound of the estimation error in each iteration is established. A thorough analysis is conducted to quantify the asymptotic behavior of the proposed method, which shows that it can significantly reduce the estimation error and also provides guideline for network merging under various scenarios. We further demonstrate the advantage of the proposed method through extensive numerical experiments on synthetic datasets and a militarized interstate dispute dataset. 
\end{abstract}	



\doublespacing

\section{Introduction}\label{Sec:intro}

Longitudinal network, also known as temporal network or continuous-time dynamic network,
consists of a sequence of temporal edges among multiple nodes, where the temporal edges may be observed between each node pair in real time \citep{holme2012temporal}. It provides a flexible framework for modeling dynamic interactions between multiple objects and how network structure evolves over time \citep{aggarwal2014evolutionary}. For instances, in online social platform such as Facebook, users send likes to the posts of their friends recurrently at different time \citep{perry2003social, snijders2010maximum}; in international politics, countries may have conflict with others at one time but become allies at others \citep{cranmer2011inferential, kinne2013network}. Similar longitudinal networks have also been frequently encountered in biological science \citep{voytek2015dynamic, avena2018communication} and ecological science \citep{ulanowicz2004quantitative, de2005dynamic}.

One of the key challenges in estimating longitudinal network resides in its scarce temporal edges, as the interactions between node pairs are instantaneous and come in a streaming fashion \citep{holme2012temporal}, and thus the observed network at each given time point can be extremely sparse. This makes longitudinal network substantially different from discrete-time dynamic network  \citep{kim2018review}, where multiple snapshots of networks are collected each with much more observed edges. In literature, various methods have been proposed for discrete-time dynamic network, such as Markov chain based methods \citep{hanneke2010discrete, sewell2015latent, sewell2016latent, matias2017statistical}, Markov process based methods \citep{snijders2010maximum, snijders2017stochastic} and tensor factorization methods \citep{lyu2023latent, han2022optimal}. Whereas the former two assume that the discrete-time dynamic network is generated from some Markov chain or Markov process, tensor factorization methods treat the discrete-time dynamic network as an order-3 tensor and often require relatively dense network snapshots. 

 
To circumvent the difficulty of severe under-sampling in longitudinal network, a common but rather ad-hoc approach is to merge longitudinal network into a multi-layer network based on equally spaced time intervals \citep{huang2023spectral}. Such an overly simplified network merging scheme completely ignores the fact that network structure may change differently during different time periods. Thus, it may introduce unnecessary estimation bias when network structure changes rapidly or incur large estimation variance when network structure stays unchanged for a long period. These negative impacts are yet neglected in literature, even though this ad-hoc network merging scheme has been widely employed to pre-process longitudinal networks in practice. Furthermore, some recent attempts were made from the perspective of survival and event history analysis \citep{vu2011continuous, vu2011dynamic, perry2013point, sit2021event}, with a keen focus on inference of the dependence of the temporal edge on some additional covariates. Some other recent works \citep{matias2018semiparametric, soliman2022multivariate}  extend the stochastic block model to detect time-invariant communities in longitudinal network.


In this paper, we propose an efficient estimation method for longitudinal network, leveraging strengths of adaptive network merging, tensor decomposition and point process. Specifically, we introduce a two-step procedure based on regularized maximum likelihood estimate to estimate the underlying tensor for the longitudinal network.  
The initial step merges the longitudinal network with some small intervals, leading to an initial estimate of the embeddings of the underlying tensor. We then adaptively merge adjacent small intervals with similar estimated temporal embedding vectors, and re-estimate the underlying tensor based on the adaptively merged intervals. A projected gradient descent algorithm is provided to facilitate estimation, as well as an information criteria for choosing the number of intervals.
A thorough theoretical analysis is conducted for the proposed estimation procedure. We first establish a general tensor estimation error bound based on a generic partition in each iteration of the projected gradient descent algorithm. The established error bound is tighter than most of the existing results in literature \citep{han2022optimal}, where the related empirical process is associated with a smaller parameter space with additional incoherence conditions. This tighter bound enables us to derive the error bound for the tensor estimate based on equally spaced intervals, which consists of an interesting bias-variance tradeoff governed by the number of small intervals and leads to faster convergence rate than that in \cite{han2022optimal} and  \cite{cai2023generalized}. More importantly, the derived error bound does not require the strong intensity condition as required in \cite{han2022optimal} and  \cite{cai2023generalized}, which, to the best of our knowledge, is the first Poisson tensor estimation error bound in both medium and weak intensity regimes. Furthermore, it is shown that the tensor estimation error, including the estimation bias and variance, can be further reduced by adaptively merging intervals, which also provides guidelines for network merging under various scenarios. The advantage of the proposed method over other existing competitors is demonstrated in extensive numerical experiments on synthetic longitudinal networks. The proposed method is also applied to analyze a militarized interstate dispute dataset, where not only the prediction accuracy increases substantially, but the adaptively merged intervals also lead to clear and meaningful interpretation.

The main contributions of  this paper are three-fold. First, we establish an upper bound for the tensor estimation error for longitudinal networks under a generic partition. By assuming additional incoherence conditions to the index set of the empirical process, our result is more powerful than the existing theoretical results. Second, we establish an upper bound for Poisson tensor estimation error under a more complete intensity regime, especially under weak and medium intensity regimes. This is a new theoretical result which has not been established in the existing literature of tensor estimation, and is of great importance in determining the best partition scheme for longitudinal network merging. Third, we propose an adaptive merging scheme for estimating the longitudinal network, and establish the upper bound for tensor estimation error. Further, we give a theoretical guideline for optimal network merging under different scenarios. It is shown that the error rates under the adaptive merging scheme are smaller than those of the equally spaced merging scheme in most scenarios.

The rest of the paper is organized as follows. Section 2 first presents the two-step estimation procedure for longitudinal network, and then propose a regularized maximum likelihood estimator based on Poisson process.
Section 3 provides the details of the computation algorithm. Section 4 establishes the error bound for the proposed method. Numerical experiments on synthetic and real-life networks are contained in Section 5. Section 6 concludes the paper with a brief discussion, and technical proofs, necessary lemmas and more numerical results are provided in the Appendix and a separate Supplementary File.

{\bf Notations.} Before moving to Section 2, we introduce some notations and preliminaries for tensor decomposition. 
For any $n\geq r$, let $\OOO_{n,r} = \{\UU\in\R^{n\times r}:\UU^\top\UU=\II_r\}$ and denote $\OOO_r = \OOO_{r,r}$. For a matrix $\UU$, let $\UU_{[i,]},\UU_{[,r]}$ and $(\UU)_{ir}$ denote the $i$-th row, $r$-th column and element $(i,r)$ of $\UU$, respectively. Let $\|\UU\|_2,\|\UU\|_F$ denote its spectral and Frobenius norm, and $\|\UU\|_{2\to\infty} = \max_{i}\|\UU_{[i,]}\|$. 
For any order-3 tensor $\MMM\in\R^{n_1\times n_2\times n_3}$, let $\MMM_{[i,,]},\MMM_{[,j,]},\MMM_{[,,k]}$
and $(\MMM)_{ijk}$ denote the $i$-th horizontal slices, $j$-th lateral slices, $k$-th frontal slices and element $(i,j,k)$ of $\MMM$, respectively. 
Let $\Psi_k(\MMM)\in\R^{n_k\times n_{-k}}$ be the mode-$k$ unfolding of $\MMM$, where $n_{-k} = n_1n_2n_3/n_k$ for $k=1,2,3$. Specifically, $$
\Psi_k(\MMM)\in\R^{n_k\times n_{-k}},~\text{where}~[\Psi_k(\MMM)]_{i_k,i_{k+1}+n_{k+1}(i_{k+2}-1)} = \MMM_{i_1i_2i_3},
$$ where $k+1$ and $k+2$ are obtained modulo 3. We denote $\rank(\MMM)\leq(r_1,r_2,r_3)$ if $\MMM$ admits the decomposition $\MMM = \SSSS\times_1\UU\times_2\VV\times_3\WW =:[\SSSS;\UU,\VV,\WW]$ for some $\SSSS\in\R^{r_1\times r_2\times r_3}$, $\UU\in\R^{n_1\times r_1}$, $\VV\in\R^{n_2\times r_2}$ and $\WW\in\R^{n_3\times r_3}$. For any order-3 tensor $\MMM$ with $\rank(\MMM)\leq(r_1,r_2,r_3)$, define $$
\begin{aligned}
	\overline\lambda(\MMM) &= \max\left\{\|\Psi_1(\MMM)\|_2,\|\Psi_2(\MMM)\|_2,\|\Psi_3(\MMM)\|_2\right\},\\
	\underline\lambda(\MMM) &= \min\left\{\sigma_{r_1}(\Psi_1(\MMM)),\sigma_{r_2}(\Psi_2(\MMM)),\sigma_{r_3}(\Psi_3(\MMM))\right\},
\end{aligned}
$$ where $\sigma_{r}(\MM)$ denote the $r$-th largest singular value of matrix $\MM$. Let $\|\MMM\|_F = \sqrt{\sum_{i,j,k}m_{ijk}^2}$ be the Frobenius norm of $\MMM$. Throughout the paper, we use $c,C,\epsilon$ and $\kappa$ to denote positive constants whose values may vary according to context. For an integer $m$, let $[m]$ denote the set $\{1,...,m\}$. For two number $a$ and $b$, let $a\wedge b = \min(a, b)$.
For two nonnegative sequences $a_{n}$ and $b_{n}$, let $a_{n}\preceq b_{n}$ and $a_{n}\prec b_{n}$ denote $a_{n} = O(b_{n})$ and $a_{n} = o(b_{n})$, respectively. Denote $a_{n}\asymp b_{n}$ if $a_{n}\preceq b_{n}$ and $b_{n}\preceq a_{n}$. Further, $a_{n}\preceq_P b_{n}$ means that there exists a positive constant $c$ such that $\Pr(a_{n}\geq cb_{n})\to0$ as $n$ diverges.

\section{Proposed method}

\subsection{Poisson point process and tensor factorization}
Consider a bipartite longitudinal network with $n_1$ out-nodes and $n_2$ in-nodes on a given time interval $[0,T)$, where $n_1$ and $n_2$ are not necessarily equal. Let  ${\cal E} = \{(i_m,j_m,t_m):m=1,...,M\}$ denote the set of all observed directed edges, where the triplet $(i,j,t)$ denotes the occurrence of a temporal edge at time $t$ pointing from out-node $i$ to in-node $j$. Note that temporal edge is instantaneous and appears at only one single time point. Let $y_{ij}(\cdot)$ be the point process that counts the number of directed edges out-node $i$ sends to in-node $j$ during $[0,T)$. Particularly, out-node $i$ sends a directed edge to in-node $j$ at time $t$ if and only if $dy_{ij}(t) = 1$. 
For each node pair $(i,j)$, suppose the intensity of $y_{ij}(t)$ is governed by some underlying propensity $\theta_{ij}(t)$. The larger $\theta_{ij}(t)$ is, the more likely out-node $i$ will send a directed edge to in-node $j$ during $[t,t+dt)$. More specifically, given $\TTT = \{\TTT(t) = (\theta_{ij}(t))_{n_1 \times n_2}\}_{t\in[0,T)}$, we assume that $y_{ij}(\cdot)$'s are mutually independent Poisson processes such that
\begin{equation}\label{eq:model}
\EEE ( dy_{ij}(t)\mid \theta_{ij}(t) ) = \lambda_0 e^{\theta_{ij}(t)}dt
\end{equation} 
where $\lambda_0>0$ is the baseline intensity. The log-likelihood function of $\{y_{ij}(t)\}_{1\leq i\leq n_1,1\leq j\leq n_2}$ can become
\begin{equation}\label{eq:poisson like}
	\begin{aligned}
		l(\TTT) = \sum_{i=1}^{n_1}\sum_{j=1}^{n_2} \left\{ \sum_{t\in\TT_{ij}} \log \lambda_{ij}(t) - \int_{0}^T\lambda_{ij}(s)ds \right\},
	\end{aligned}
\end{equation} 
where $\lambda_{ij}(t) = \lambda_0\exp(\theta_{ij}(t))$. Note that $\lambda_0$ is fixed throughout the paper, but it could also be varying with $t$, which may require more involved treatment.

Suppose $\TTT(t)$ admits a low rank structure so that
\begin{equation}\label{eq:low rank}
	\theta_{ij}(t) = \SSSS\times_1\uu_i^\top\times_2\vv_j^\top\times_3\ww(t)^\top,
\end{equation}
where $\times_s$ denotes the mode-$s$ product for $s \in [3]$, $\SSSS\in\R^{r_1\times r_2\times r_3}$ is an order-3 core tensor, and each out-node $i$, in-node $j$ and time $t$ are embedded as low-dimensional vectors $\uu_i\in\R^{r_1},\vv_j\in\R^{r_2}$ and $\ww(t)\in\R^{r_3}$, respectively. It is clear that the time-invariant network structure is captured by the network embedding vectors $\uu$ and $\vv$, while the temporal structure is captured by the temporal embedding vector $\ww(t)$. Such a network embedding model has been widely employed for network data analysis \citep{hoff2002latent, lyu2023latent,zhang2021directed,zhen2023community}, which embeds the unstructured network in a low-dimensional Euclidean space to facilitate the subsequent analysis. It is also related to the random dot product graph model \citep{athreya2017statistical, rubin2022statistical}.



\subsection{Adaptive merging}

Let ${\cal G}_t = \{(i,j):(i,j,t)\in{\cal E}\}$ as the observed network at time $t$, $\TT_{ij} = \{t\in[0,T):(i,j,t)\in{\cal E}\}$ as the time stamps for directed edges $(i,j)$. Since the directed edges in ${\cal E}$ are observed in real time, ${\cal G}_{t}$ can be extremely sparse and may consist of even only one observed edge, which casts great challenge for estimating the longitudinal network. To circumvent the difficulty of severe under-sampling, we propose to embed the longitudinal network by adaptively merging ${\cal G}_t$ into relatively dense networks based on their temporal structures,  which leads to a substantially improved estimation of the longitudinal network.

We first split the time window $[0,T)$ into $L$ equally spaced small intervals with endpoints $\{\delta_l\}_{l=1}^L$, where $\delta_l = l \Delta_{\ddd}$, $\delta_0 = 0$, and each interval $[\delta_{l-1},\delta_l)$ is of width $\Delta_{\ddd} = T/L$. When $\Delta_{\ddd}$ is sufficiently small, it is expected that $\TTT(t)$ shall be roughly constant within each time interval. As a direct consequence, $\TTT(t)$ can be estimated by a low rank order-3 tensor $\MMM\in\R^{n_1\times n_2\times L}$, which admits a Tucker decomposition with rank $(r_1,r_2,r_3)$, 
$$
\MMM = \SSSS \times_1 \UU \times_2 \VV \times_3 \WW,
$$ 
with $\uu_i,\vv_j$ and $\ww_l$ being the corresponding rows of $\UU,\VV$ and $\WW$, respectively.  Let $\ddd = (\delta_1,...,\delta_{L})^\top$ and $\YYY_{\ddd}\in\R^{n_1\times n_2\times L}$ with $(\YYY_{\ddd})_{ijl} = |\TT_{ij}\cap[\delta_{l-1},\delta_l)|$ representing the number of temporal edges in each small interval. An initial estimate $\widehat\MMM_{\ddd} = [\widehat\SSSS_{\ddd}; \widehat\UU_{\ddd}, \widehat\VV_{\ddd}, \widehat\WW_{\ddd}]$ can be obtained by minimizing certain distance measure between $\MMM$ and $\YYY_{\ddd}$, to be specified in Section~\ref{subsec:like}.

Once the initial estimate $\widehat\WW_{\ddd} = (\widehat\ww_{1,\ddd},...,\widehat\ww_{L,\ddd})^\top$ is obtained, define 
\begin{equation}\label{eq:normalize}
\widetilde\WW_{\ddd} =(\widetilde\ww_{1,\ddd},...,\widetilde\ww_{L,\ddd})^\top= \sqrt{L}\widehat\WW_{\ddd}((\widehat\WW_{\ddd})^\top\widehat\WW_{\ddd})^{-\frac{1}{2}},
\end{equation}
where  $(\widehat\WW_{\ddd})^\top\widehat\WW_{\ddd}$ is invertible with high probability as to be shown in the proof of Theorem~\ref{thm:cluster}. This is actually a normalization step to facilitate technical analysis. Though consistent, the estimation variance of $\widetilde\WW_{\ddd}$ can be exceedingly large when $\Delta_{\ddd}$ is too small. We then propose to merge adjacent small intervals with similar temporal embedding vectors $\widetilde\ww_{l,\ddd}$, so as to shrink the estimation variance without compromising the estimation bias. 

Let $\PPP = \{\PPP_1,...,\PPP_K\}$ denote the adaptively merged intervals, where for any $l_1\in \PPP_{k_1}$ and $l_2\in \PPP_{k_2}$, it holds that $l_1<l_2$ if $k_1<k_2$. Then, it can be estimated as
\begin{equation}\label{eq:cluster}
\widehat\PPP = \argmin_{\PPP} \sum_{k=1}^K\sum_{l\in\PPP_k} \|\widetilde\ww_{l,\ddd}-\mm_k\|^2,
\end{equation} 
where $\mm_k = |\PPP_k|^{-1} \sum_{l\in\PPP_k}\widetilde\ww_{l,\ddd}$. Note that \eqref{eq:cluster} is equivalent to seeking change points in the sequence $(\widetilde\ww_{1,\ddd},...,\widetilde\ww_{L,\ddd})$, and thus can be efficiently solved by multiple change point detection algorithm \citep{hao2013multiple, niu2016multiple}.
Further, define $\widehat\eta_k = \DDD_{\ddd}\max \widehat\PPP_k$, and thus $\widehat\hhh = (\widehat\eta_1,...,\widehat\eta_K)^\top$ consists of the estimated endpoints of $K$ adaptively merged intervals.  Denote $\YYY_{\widehat\hhh}\in\R^{n_1\times n_2\times K}$ with $(\YYY_{\widehat\hhh})_{ijk} = |\TT_{ij}\cap[\widehat\eta_{k-1},\widehat\eta_k)|$ with $\widehat\eta_0 = 0$, the final estimate  $\widehat\MMM_{\widehat\hhh}$ is then obtained by minimizing the distance measure between $\MMM$ and $\YYY_{\widehat \hhh}$.

\subsection{Regularized likelihood estimation}\label{subsec:like}

Let $\tttt=(\tau_1,\ldots,\tau_{n_3})^\top$ denote a generic partition of $[0,T)$ with $0 =\tau_0 < \tau_1 <...<\tau_{n_3} = T$. Particularly, $\tttt$ could be the equally spaced intervals $\ddd$ for the initial estimate or the adaptively merged intervals $\widehat\hhh$ for the final estimate, and $n_3$ could be $L$ or $K$, correspondingly. For any $\MMM\in\R^{n_1\times n_2\times n_3}$, we define
\begin{equation}\label{eq:like gene}
\begin{aligned}
l(\MMM;\tttt) = \sum_{i=1}^{n_1}\sum_{j=1}^{n_2}\sum_{l=1}^{n_3}  \big \{ m_{ijl}\left|\TT_{ij}\cap[\tau_{l-1},\tau_l)\right| - e^{m_{ijl}}\lambda_0 (\tau_{l}-\tau_{l-1}) \big \}.
\end{aligned}
\end{equation}
Note that if $\TTT(t)$ is roughly constant in each interval, we consider the regularized formulation,
\begin{equation}\label{eq:opti gene}
\begin{aligned}
(\widehat\SSSS_{\tttt},\widehat\UU_{\tttt},\widehat\VV_{\tttt},\widehat\WW_{\tttt}) =& \argmin_{\SSSS,\UU,\VV,\WW}  \  \left\{ -l(\MMM;\tttt) + \gamma\JJJ_{\tttt}(\UU,\VV,\WW)\right\},
\end{aligned}
\end{equation}
where $\gamma$ is the tuning parameter, $\JJJ_{\tttt}(\UU,\VV,\WW)$ is the regularization term which takes the form $$
\JJJ_{\tttt}(\UU,\VV,\WW) = \frac{1}{4}\left\{\Big\|\frac{1}{n_1}\UU^\top\UU - \II_{r_1}\Big\|_F^2 + \Big\|\frac{1}{n_2}\VV^\top\VV - \II_{r_2}\Big\|_F^2 + \Big\|\frac{1}{n_3}\WW^\top\WW - \II_{r_3}\Big\|_F^2\right\},
$$ 
encouraging the orthogonality among columns in $\UU,\VV$ and $\WW$. A similar regularization term has also been employed in \cite{han2022optimal}, which involves some additional tuning parameter and thus requires more computational efforts.

\section{Computation}\label{sec:compute}

Define $\CCC_{\SSSS} = \{\SSSS\in\R^{r_1\times r_2\times r_3}:\|\SSSS\|_{F}\leq c_{\SSSS}\},~\CCC_{\UU} = \{\UU\in\R^{n_1\times r_1}:\|\UU\|_{2\to\infty}\leq c_1\},~\CCC_{\VV} = \{\VV\in\R^{n_2\times r_2}:\|\VV\|_{2\to\infty}\leq c_2\}$, and $
\CCC_{\WW} = \{\WW\in\R^{n_3\times r_3}:\|\WW\|_{2\to\infty}\leq c_3\}$, where $c_{\SSSS},c_1,c_2$ and $c_3$ are constants. 
Here $n_3$ could be $L$ and $K$ and with a little abuse of notation, we use a generic $\CCC_{\WW}$.
For any convex set $\CCC$, denote $\PPP_{\CCC}$ to be the projection operator onto $\CCC$.

We develop an efficient projected gradient descent (PGD) updating algorithm to solve the optimization task in \eqref{eq:opti gene}. Choose an initializer $(\SSSS^{(0)}_{\tttt},\UU^{(0)}_{\tttt},\VV^{(0)}_{\tttt},\WW^{(0)}_{\tttt})$ such that $\SSSS^{(0)}_{\tttt}\in\CCC_{\SSSS},~\UU^{(0)}_{\tttt}\in\CCC_{\UU},~\VV^{(0)}_{\tttt}\in\CCC_{\VV}$ and $\WW^{(0)}_{\tttt}\in\CCC_{\WW}$, with ${\UU^{(0)}_{\tttt}}^\top{\UU^{(0)}_{\tttt}} = n_1\II_{r_1},~{\VV^{(0)}_{\tttt}}^\top{\VV^{(0)}_{\tttt}} = n_2\II_{r_2}$ and ${\WW^{(0)}_{\tttt}}^\top{\WW^{(0)}_{\tttt}} = n_3\II_{r_3}$. 
Given $(\SSSS^{(r)}_{\tttt},\UU^{(r)}_{\tttt},\VV^{(r)}_{\tttt},\WW^{(r)}_{\tttt})$ and $\MMM^{(r)}_{\tttt} = [\SSSS^{(r)}_{\tttt}; \UU^{(r)}_{\tttt},\VV^{(r)}_{\tttt},\WW^{(r)}_{\tttt}]$,
we implement the following updating scheme with step size $\zeta$: 
{\small
\begin{equation}\label{eq:update gene}
\begin{aligned}
\UU^{(r+1)}_{\tttt} &= \PPP_{\CCC_{\UU}}\left\{\UU^{(r)}_{\tttt} + \zeta\left[ n_1\frac{\partial l(\MMM^{(r)}_{\tttt};\tttt)}{\partial\UU} - \gamma\UU^{(r)}_{\tttt}\left(\frac{1}{n_1}{\UU^{(r)}_{\tttt}}^\top\UU^{(r)}_{\tttt} - \II_{r_1}\right)  \right]\right\}; \\
\VV^{(r+1)}_{\tttt} &= \PPP_{\CCC_{\VV}}\left\{\VV^{(r)}_{\tttt} + \zeta\left[ n_2\frac{\partial l(\MMM^{(r)}_{\tttt};\tttt)}{\partial\VV} - \gamma\VV^{(r)}_{\tttt}\left(\frac{1}{n_2}{\VV^{(r)}_{\tttt}}^\top\VV^{(r)}_{\tttt} - \II_{r_2}\right)  \right]\right\};\\
\WW^{(r+1)}_{\tttt} &= \PPP_{\CCC_{\WW}}\left\{\WW^{(r)}_{\tttt} + \zeta\left[ n_3\frac{\partial l(\MMM^{(r)}_{\tttt};\tttt)}{\partial\WW} - \gamma\WW^{(r)}_{\tttt}\left(\frac{1}{n_3}{\WW^{(r)}}_{\tttt}^\top\WW^{(r)}_{\tttt} - \II_{r_3}\right)  \right]\right\};\\
\SSSS^{(r+1)}_{\tttt} &= \PPP_{\CCC_{\SSSS}}\left\{\SSSS^{(r)}_{\tttt} + \zeta\frac{\partial l(\MMM^{(r)}_{\tttt};\tttt)}{\partial\SSSS}\right\},
\end{aligned}
\end{equation}
}
and let $\MMM^{(r+1)}_{\tttt} = [\SSSS^{(r+1)}_{\tttt}; \UU^{(r+1)}_{\tttt},\VV^{(r+1)}_{\tttt},\WW^{(r+1)}_{\tttt}]$. We repeat the above updating scheme for a relative large number of iterations, say $R$, and 
let $(\widehat\SSSS_{\tttt},\widehat\UU_{\tttt},\widehat\VV_{\tttt},\widehat\WW_{\tttt}) = (\SSSS_{\tttt}^{(R)},\UU_{\tttt}^{(R)},\VV_{\tttt}^{(R)},\WW_{\tttt}^{(R)})$ be the initial estimation. 

\begin{remark}
We point out that the updating scheme in \eqref{eq:update gene} differs from the standard projected gradient descent update, as different step sizes are used for updating different variables. Specifically, the step sizes for updating $\UU^{(r)}_{\tttt},\VV^{(r)}_{\tttt},~\WW^{(r)}_{\tttt},\SSSS^{(r)}_{\tttt}$ are $n_1\zeta,n_2\zeta,n_3\zeta$ and $\zeta$, respectively. This is the key difference from the algorithm in \cite{han2022optimal}, which is also the reason that we do not require additional tuning parameter in $\JJJ_{\tttt}(\UU,\VV,\WW)$ and $\JJJ_{\hhh}(\UU,\VV,\WW)$. As will be shown in Theorem~\ref{thm:tensor1} and \ref{thm:tensor2}, $\zeta$ is chosen as $\frac{c}{n_1n_2T}$.
\end{remark}

It remains to determine the number of merged interval $K$ in \eqref{eq:cluster}. In particular, we set 
\begin{equation}\label{eq:IC}
	\widehat{K} = \argmin_{S} \big\{\min_{\PPP} \LLL(\PPP;S) +  \nu_{nT} S \big \},
\end{equation} 
where $\PPP = \{\PPP_1,...,\PPP_S\}$ is an ordered partition of $[L]$, $\nu_{nT}$ is a quantity to be specified in Theorem \ref{thm:cluster}, and
\begin{equation}\label{eq:cluster loss}
\LLL(\PPP;S) = \frac{1}{L}\sum_{s=1}^S\sum_{l\in\PPP_s} \|\widetilde\ww_{l,\ddd}-\mm_s\|^2,
\end{equation}
with $\mm_s = |\PPP_s|^{-1} \sum_{l\in\PPP_s}\widetilde\ww_{l,\ddd}$. More importantly, $\widehat K$ is a consistent estimator of $K$ as to be shown in Theorem \ref{thm:cluster}, which can be technically more involved than estimating the number of change points in $(\widetilde\ww_{1,\ddd},...,\widetilde\ww_{L,\ddd})$, due to the mutual dependence among $\widetilde\ww_{1,\ddd},...,\widetilde\ww_{L,\ddd}$. 

\begin{figure}[!htb]
\caption{Flowchart for the estimation procedure, where $n = \max\{n_1,n_2\}$ and the logarithmic factors are suppressed.}
\label{fig:algo}
\begin{tikzpicture}[node distance=2cm]
\node (net) [startstop] {${\cal E}$};
\node (delta) [startstop, right of=net, xshift=2.2cm] {$(\ddd,\YYY_{\ddd})$};
\node (est1) [startstop, right of=delta, xshift=0.2cm] {$\widehat\MMM_{\ddd}$};
\node (dec) [decision, right of=est1, xshift=0.1cm] {{\tiny $T\succ n^{\frac{2}{3}}$}};
\node (eta) [startstop, right of=dec, xshift=2cm] {$(\widehat\hhh,\YYY_{\widehat\hhh})$};
\node (est2) [startstop, right of=eta, xshift=0.2cm] {$\widehat\MMM_{\widehat\hhh}$};

\draw [arrow] (net) -- node[anchor=south] {{\tiny Equal spaced merging}} 
				node[anchor=north] {{\tiny $L\asymp n\sqrt{T}$ }} (delta);
\draw [arrow] (delta) -- node[anchor=south] {{\tiny PGD}} (est1);
\draw [arrow] (est1) -- (dec);
\draw [arrow] (dec) -- node[anchor=south] {{\tiny Yes}} 
             			  node[anchor=north] {{\tiny Adaptive merging}}	(eta);
\draw [arrow] (eta) -- node[anchor=south] {{\tiny PGD}} (est2);
\end{tikzpicture}
\end{figure}
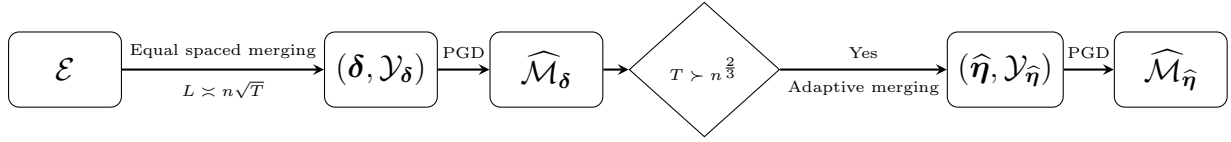

Figure~\ref{fig:algo} gives a visual illustration for the proposed procedure, and Algorithm~\ref{algo:code} further gives more detailed implementations. 
The Tucker ranks $(r_1, r_2, r_3)$ in the requirement of Algorithm~\ref{algo:code} could be selected based on $\YYY_{\ddd}$ in the same way as Han et al. (2022). 

\begin{algorithm}
\caption{Estimating longitudinal networks via adaptive merging}\label{algo:code}
\begin{algorithmic}[1]
\Require Temporal edges ${\cal E} = \{(i_m,j_m,t_m)\}_{m=1}^M$ and $(n_1,n_2, T)$, Tucker ranks $(r_1, r_2, r_3)$, baseline intensity $\lambda_0>0$, step size constant $c>0$, constraint parameters $(c_{\SSSS},c_1,c_2,c_3)$
\vspace{1mm}
\State Determine $L$ and the equal spaced partition $\ddd$ according to Table~\ref{tab:res}; 
\State Formulate response tensor $\YYY_{\ddd}$ based on ${\cal E}$;
\State Perform \eqref{eq:update gene} based on $(\ddd,\YYY_{\ddd})$ to obtain $\widehat\MMM_{\ddd} = [\widehat\SSSS_{\ddd}; \widehat\UU_{\ddd}, \widehat\VV_{\ddd}, \widehat\WW_{\ddd}]$;
\If{$T\preceq n^{\frac{2}{3}}\log^{1+\frac{2}{3}\epsilon}(nT)$} \Comment{See Section 4 for more details.}
	\State \Return $\widehat\MMM_{\ddd} = [\widehat\SSSS_{\ddd}; \widehat\UU_{\ddd}, \widehat\VV_{\ddd}, \widehat\WW_{\ddd}]$.   
\Else
	\State Determine $K$ based on \eqref{eq:IC}, and obtain $\widehat\hhh$ based on \eqref{eq:cluster};
	\State Formulate response tensor $\YYY_{\widehat\hhh}$;
	\State Perform \eqref{eq:update gene} based on $(\widehat\hhh, \YYY_{\widehat\hhh})$ to obtain $\widehat\MMM_{\widehat\hhh} = [\widehat\SSSS_{\widehat\hhh}; \widehat\UU_{\widehat\hhh}, \widehat\VV_{\widehat\hhh}, \widehat\WW_{\widehat\hhh}]$;  
	\State\Return $\widehat\MMM_{\widehat\hhh} = [\widehat\SSSS_{\widehat\hhh}; \widehat\UU_{\widehat\hhh}, \widehat\VV_{\widehat\hhh}, \widehat\WW_{\widehat\hhh}]$.
\EndIf
\end{algorithmic}
\end{algorithm}

\section{Theory}\label{sec:theory}

Suppose the longitudinal network ${\cal G}_t$ is generated with $\TTT^*(t) = \SSSS^* \times_1 \UU^* \times_2 \VV^* \times_3 \ww^*(t)$, where $\rank(\Psi_s(\SSSS^*)) = r_s$ for $s=1,2,3$,  ${\UU^*}^\top\UU^* = n_1\II_{r_1}$, ${\VV^*}^\top\VV^* = n_2\II_{r_2}$ and $\int_{0}^T \ww^*(t)\ww^*(t)^\top dt = T\II_{r_3}$. Further, suppose $\ww^*(t)$ is a piecewise constant function of $t$ in that $\ww^*(t) = \ww^*_{k,\hhh}$ for $t\in[\eta_{k-1},\eta_{k})$, where $\hhh = (\eta_1,...,\eta_{K_0})^\top$ with $0=\eta_0<\eta_1<...<\eta_{K_0}=T$. Let $\WW^*_{\hhh}\in\R^{{K_0}\times r_3}$ with $(\WW^*_{\hhh})_{[k,]} = \ww^*_{k,\hhh}$,  and $\MMM^*_{\hhh} = [\SSSS^*;\UU^*,\VV^*,\WW^*_{\hhh}]$. 


\subsection{A new error bound for the PGD algorithm}\label{subsec:key}

We first derive the upper bound for the tensor estimation error in each iteration of the PGD algorithm \eqref{eq:update gene}. Let $\tttt = (\tau_1,...,\tau_{n_3})^\top\in\R^{n_3}$ with
$0 =\tau_0 < \tau_1 <...<\tau_{n_3} = T$ be a generic partition of $[0,T)$, which could be $\ddd$ or $\widehat\hhh$. Recall that 
\begin{align}\label{eq:indexset}
l(\MMM;\tttt) &= \sum_{i=1}^{n_1}\sum_{j=1}^{n_2}\sum_{k=1}^{n_3}  \left\{ m_{ijk}\big|\TT_{ij}\cap[\tau_{k-1},\tau_k)\big| - e^{m_{ijk}}\lambda_0(\tau_k-\tau_{k-1}) \right\},~~\text{and define} \nonumber \\
\widetilde\CCC_{\MMM,\tttt}  &=  \Big\{ \MMM = [\SSSS;\UU,\VV,\WW]:~\SSSS\in\R^{r_1\times r_2\times r_3},\UU\in\R^{n_1\times r_1},\VV\in\R^{n_2\times r_2},\WW\in\R^{n_{3}\times r_3}, \nonumber \\
&\|\SSSS\|_F=\|\UU\|_F=\|\VV\|_F=\|\WW\|_F=1,
~\text{and at least two of the followings hold:}\nonumber\\
&\|\UU\|_{2\to\infty}\leq 2c_1n_1^{-1/2},~~\|\VV\|_{2\to\infty}\leq2c_2n_2^{-1/2},~~\|\WW\|_{2\to\infty}\leq2c_3n_3^{-1/2}\Big\}.
\end{align}

Given $\tttt$, for any tensor $\overline\MMM\in\R^{n_1\times n_2\times n_3}$, define $\overline\MMM_{\tttt}(t) = (\overline\MMM)_{[,,l]}$ for any $t\in[\tau_{l-1},\tau_{l})$, and 
$$
\xi_{\tttt}(\overline\MMM) = \sup_{\MMM\in\widetilde\CCC_{\MMM,\tttt}} \big| \langle \nabla l(\overline\MMM;\tttt),\MMM \rangle \big|,
$$ 
which essentially quantifies the difference between $\overline\MMM$ and a stationary point of $l(\cdot;\tttt)$ \citep{han2022optimal}. Actually, $\xi_{\tttt}(\overline\MMM)$ also measures the amplitude of $\nabla l(\overline\MMM;\tttt)$ projected onto the manifold of tensors with ranks $(r_1,r_2,r_3)$ under the incoherence conditions. It is important to remark that the incoherence conditions in $\widetilde\CCC_{\MMM,\tttt}$ are the key factors to relax the strong intensity condition as required in \cite{han2022optimal}. Furthermore, note that
\begin{equation}\label{eq:xi exp}
\begin{aligned}
\xi_{\tttt}(\overline\MMM)
\leq& \sup_{\MMM\in\widetilde\CCC_{\MMM,\tttt}} \big| \langle \nabla l(\overline\MMM;\tttt) - \EEE\nabla l(\overline\MMM;\tttt),\MMM \rangle \big| + \sup_{\MMM\in\widetilde\CCC_{\MMM,\tttt}} \big| \langle \EEE\nabla l(\overline\MMM;\tttt),\MMM \rangle \big| \\
=& I_1 + I_2,
\end{aligned}
\end{equation} 
where $I_1$ characterize the amplitude of the statistical noise, and $I_2$ quantifies the bias between $\overline\MMM_{\tttt}(t)$ and $\TTT^*(t)$. To see this, if $l(;\tttt)$ is deterministic and $\overline\MMM$ is a stationary point, $I_1=0$; and if $\overline\MMM_{\tttt}(t) = \TTT^*(t)$, 
\begin{align*}
\EEE\nabla l(\overline\MMM;\tttt)_{[i,j,k]} & = \EEE\left|\TT_{ij}\cap[\tau_{k-1},\tau_k)\right| - e^{\overline m_{ijk}}\lambda_0 (\tau_{k}-\tau_{k-1}) \\ 
& = \lambda_0(e^{\theta_{ij}^*(\tau_{k-1})}-e^{\overline m_{ijk}})(\tau_{k}-\tau_{k-1}) = 0~\text{for any}~(i,j,k),
\end{align*}
and thus $I_2 = 0$, whereas $I_2$ would be substantially larger than 0 if $\overline\MMM_{\tttt}(t)$ differs from $\TTT^*(t)$. 

Note that if $\tttt$ is not a superset of $\hhh$, there exists no $\overline\MMM$ such that $\overline\MMM_{\tttt}(t)\neq\TTT^*(t)$ for any $t$. Therefore, for any pre-specified tensor $\overline\MMM$ with bounded $\xi_{\tttt}(\overline\MMM)$, its estimation error by $\MMM_{\tttt}^{(R)}$ is established in Theorem~\ref{thm:key}.

\begin{theorem}\label{thm:key}
Let $\overline\MMM = [\overline\SSSS;\overline\UU,\overline\VV,\overline\WW]\in\R^{n_1\times n_2\times n_3}$ be a pre-specified order-3 tensor with $\overline\SSSS\in\CCC_{\SSSS},~\overline\UU\in\CCC_{\UU}$, $\overline\VV\in\CCC_{\VV}$, $\overline\WW\in\CCC_{\WW}$,  $\overline\UU^\top\overline\UU = n_1\II_{r_1}$, $\overline\VV^\top\overline\VV = n_2\II_{r_2}$, $\overline\WW^\top\overline\WW = n_3\II_{r_3}$, and $\underline\lambda(\overline\SSSS) \asymp \overline\lambda(\overline\SSSS)\asymp1$.


Suppose there exists a quantity $H\in(0,T]$ such that $\min_{1\leq l\leq n_3}(\tau_{l}-\tau_{l-1}) \asymp \max_{1\leq l\leq n_3}(\tau_{l}-\tau_{l-1})\asymp H$. Further suppose $\gamma \asymp n_1n_2n_3H$ and $\xi_{\tttt}(\overline\MMM) \preceq_P \sqrt{n_1n_2n_3}H$. Then there exists $c_0>0$ such that for any step size $\zeta = \frac{c}{n_1n_2n_3H}$ with $0<c<c_0$, we have 
\begin{equation}\label{eq:error}
\frac{1}{n_1n_2n_3}\|\MMM^{(R)}_{\tttt} - \overline\MMM\|_F^2   \preceq_P \frac{\xi_{\tttt}(\overline\MMM)^2}{n_1n_2n_3H^2}+c_1(1 - \kappa)^R,
\end{equation}
for some constant $0<\kappa<1$ and $c_1>0$.
\end{theorem}

The term $c(1 - \kappa)^R$ in \eqref{eq:error} is the optimization error which decays linearly with iterations. Thanks to the regularizer in  \eqref{eq:opti gene} and the restricted correlated gradient condition \citep{han2022optimal} of the log-likelihood function, $\MMM^{(R)}_{\tttt}$ will converge to a stationary point of the log-likelihood function at a linear convergence rate as $R$ grows. The upper bound in the right-hand side of \eqref{eq:error} is thus dominated by the first term. Actually, we shall choose $\overline\MMM$ to be $\MMM^*_{\ddd}$ in the initial estimate, which is specified in Theorem~\ref{thm:tensor1}, and $\MMM^*_{\hhh}$ in the final estimate. The asymptotic orders of the corresponding $\xi_{\tttt}(\overline\MMM)$ are established in Theorems~\ref{thm:tensor1} and \ref{thm:tensor2}.

It is interesting to remark that a similar upper bound for the tensor estimation error is established in \cite{han2022optimal}. Yet, Theorem \ref{thm:key} differs from \cite{han2022optimal} in that the space $\widetilde\CCC_{\MMM,\tttt}$ associated with the empirical process $\xi_{\tau}$ is reduced by requiring additional incoherence conditions that $\|\UU\|_{2\to\infty}\leq\mu_1$, $\|\VV\|_{2\to\infty}\leq\mu_2$ and $\|\WW\|_{2\to\infty}\leq\mu_3$ with $\mu_k = \sqrt{\log(n_k)/n_k}$. The incoherence conditions for $\UU,\VV,\WW$ in $\widetilde\CCC_{\MMM,\tttt}$ are the key ingredients to derive the convergence rate for the tensor estimation error in a complete regime, in contrast to the results in \cite{han2022optimal} and \cite{cai2023generalized} requiring the strong intensity condition.

The following proposition quantifies the Poisson tensor estimator error in the strong, medium and weak intensity regimes. The error bound in the strong intensity regime has been established in \cite{han2022optimal}, while the error bounds in the other two regimes are new addition to the literature. It will be shown in Sections~\ref{subsec:equal} and \ref{subsec:merge} that the derived error bound in the medium and weak intensity regimes are of great importance.

\begin{proposition}\label{prop:PoissonPCA}
Let $\YYY\in\R^{n_1\times n_2\times n_3}$ be a random tensor whose entries follow Poisson distribution with mean $I \exp(\overline\MMM)$, with $I>0$ and $\overline\MMM= [\overline\SSSS;\overline\UU,\overline\VV,\overline\WW]\in\R^{n_1\times n_2\times n_3}$ satisfying the same conditions as in Theorem~\ref{thm:key}.
For any $\MMM\in\R^{n_1\times n_2\times n_3}$, let $l(\MMM) = \sum_{i,j,l}(m_{ijl}y_{ijl} -Ie^{m_{ijl}})$, and $\widehat \MMM^{(R)}$ be the estimate after $R$ iteration of \eqref{eq:update gene}, where $l(\MMM;\tttt)$ is replaced by $l(\MMM)$. Suppose $\gamma \asymp n_1n_2n_3I$ and $\xi(\overline\MMM) := \sup_{\MMM\in\widetilde\CCC_{\MMM}} \big| \langle \nabla l(\overline\MMM),\MMM \rangle \big| \preceq_P \sqrt{n_1n_2n_3}I$, where $\widetilde\CCC_{\MMM}$ is defined the same as $\widetilde\CCC_{\MMM,\tttt}$ in \eqref{eq:indexset}. Then there exists $c_0>0$ such that for any step size $\zeta = \frac{c}{n_1n_2n_3I}$ with $0<c<c_0$, we have $$
\frac{1}{n_1n_2n_3}\|\widehat\MMM^{(R)}-\overline\MMM\|_F^2 \preceq  C(1 - \kappa)^R + 
\left\{
\begin{aligned}
&\frac{n_1+n_2+n_3}{n_1n_2n_3I},&&\mbox{if }~I\succ \log(n_1n_2n_3),\\
&\frac{(n_1+n_2+n_3)\log^{1+\epsilon}(n_1n_2n_3)}{n_1n_2n_3I},&& \mbox{if }~1\preceq I\preceq \log(n_1n_2n_3),\\
&\frac{(n_1+n_2+n_3)\log(n_1n_2n_3)}{n_1n_2n_3I},&& \mbox{if }~\frac{\log(n_1n_2n_3)}{n_1\wedge n_2\wedge n_3}\prec I\prec 1,
\end{aligned}
\right.
$$ for some constant $0<\kappa<1$ and $c_1>0$.
\end{proposition}

\begin{remark}
Thanks to the new error bound for the PGD algorithm in Theorem \ref{thm:key}, the strong intensity condition $I\succ \log(n_1n_2n_3)$ as required in \cite{han2022optimal} and \cite{cai2023generalized} can be relaxed, and similar upper bound can be obtained even when $I$ decays to $0$. To the best of our knowledge, Proposition~\ref{prop:PoissonPCA} gives the first Poisson tensor estimation error bound in both weak intensity regime with $I\prec1$ and medium intensity regime with $1\preceq I\preceq \log(n_1n_2n_3)$.
If we suppress the logarithmic factor, the error bound is essentially $\frac{n_1+n_2+n_3}{n_1n_2n_3I}$ in all regimes. 
\end{remark}


\subsection{Error analysis based on equally spaced intervals} \label{subsec:equal}

Let $n = \max\{n_1,n_2\}$ for simplicity and suppose $n_1\asymp n_2 \asymp n$. 
Define $d_{\min} = \min_{1\leq k\leq {K_0}}(\eta_k - \eta_{k-1})/T$, $d_{\max} = \max_{1\leq k\leq {K_0}}(\eta_k - \eta_{k-1})/T$ and $\DDD_{\hhh} = d_{\min}T$. Suppose $d_{\min} \asymp d_{\max}\asymp1/K_0$, which requires that the lengths of all intervals based on $\hhh$ are of the same order.
Further suppose $\|\SSSS^*\|_{F}\leq c_{\SSSS}/\max\{2,(K_0d_{\min})^{-1/2}\}$, $\|\UU^*\|_{2\to\infty}\leq c_1$,
$\|\VV^*\|_{2\to\infty}\leq c_2$ and $\sup_{t\in[0,T)}\|\ww^*(t)\|\leq c_3 / \max\{2,\sqrt{K_0d_{\max}}\}$, where $(c_{\SSSS},c_1,c_2,c_3)$ are defined in $(\CCC_{\SSSS}, \CCC_{\UU}, \CCC_{\VV}, \CCC_{\WW})$ in the beginning of Section~\ref{sec:compute}, and the different requirements for $\|\SSSS\|_F$ and $\|\ww^*(t)\|$ are due to the normalization step \eqref{eq:normalize}.
Recall that $\DDD_{\ddd} = T/L$. Theorem~\ref{thm:tensor1} establishes the tensor error bound for the initial estimate based on the equally spaced interval $\ddd$.

\begin{theorem}\label{thm:tensor1} {\bf (Initial estimate)}
Choose $\gamma \asymp n^2T$ and $\zeta = \frac{c}{n^2T}$ for some small constant $c>0$. Then, with probability approaching 1, it holds true that
$$
\frac{1}{n_1n_2L}\|\MMM^{(R)}_{\ddd} - \MMM^*_{\ddd}\|_F^2  \preceq I_{1,\ddd}+I_{2,\ddd}+I_{3,\ddd},
$$ 
where $\MMM^*_{\ddd} = [\SSSS^*;\UU^*,\VV^*,\WW^*_{\ddd}]$ with $\WW^*_{\ddd}\in\R^{L\times r_3}$ such that $(\WW^*_{\ddd})_{[l,]} = \ww^*(\delta_{l-1})$. Here $$
I_{1,\ddd} = \left\{
\begin{aligned}
&\frac{1}{nT} + \frac{L}{n^2T},&&\mbox{if }~\log(nT)\prec\DDD_{\ddd}\prec \frac{T}{K_0},\\
&\frac{\log^{1+\epsilon}(nT)}{nT} + \frac{L\log^{1+\epsilon}(nT)}{n^2T},&& \mbox{if }~1\preceq \DDD_{\ddd}\preceq \log(nT),\\
&\frac{\log(nT)}{nT} + \frac{L\log(nT)}{n^2T},&& \mbox{if }~\frac{(n+L)^2\log(nT)}{n^2L}\prec \DDD_{\ddd}\prec 1, 
\end{aligned}
\right.
$$ for any $\epsilon>0$, $I_{2,\ddd} = K_0/L$ and $I_{3,\ddd} = C(1 - \kappa)^R$ for some constants $C$ and $0<\kappa<1$.
\end{theorem}


Respectively, $I_{1,\ddd}$, $I_{2,\ddd}$ and $I_{3,\ddd}$ correspond to the estimation variance, the bias induced by network merging, and the optimization error of \eqref{eq:update gene} after $R$ iterations. If we suppress the logarithmic factor, the estimation variance $I_{1,\ddd}\asymp \frac{1}{nT}+\frac{L}{n^2T}$, which matches up with the minimax lower bound for the tensor estimation error in Poisson PCA \citep{han2022optimal}.

\begin{remark}
Given the partition $\ddd$, the problem becomes estimating the low-rank $\MMM_{\ddd}$ based on $\YYY_{\ddd}$ with $(\YYY_{\ddd})_{ijl} = |\TT_{ij}\cap[\delta_{l-1},\delta_l)|$, where $(\YYY_{\ddd})_{ijl}$ follows the Poisson distribution with intensity $\int_{(l-1)\DDD_{\ddd}}^{l\DDD_{\ddd}}\ttt_{ij}^*(t)dt \propto \DDD_{\ddd}$.  The results in \cite{han2022optimal} and \cite{cai2023generalized} require that $\DDD_{\ddd}\succ \log(nT)$, or the intensity needs to be ``strong'', whereas Theorem~\ref{thm:tensor1} still holds when $\DDD_{\ddd} \preceq \log(nT)$ or even $\DDD_{\ddd}\preceq 1$. As will be shown in Corollary~\ref{cor:tensor1} and Remark~\ref{rk:tensor1}, allowing $\DDD_{\ddd}\prec 1$ will lead to a faster convergence rate in certain scenario. We establish the upper bound in the weak and medium intensity regimes by exploiting a more delicate concentration inequality. Thanks to the additional incoherence conditions in Theorem 1, we use the Chernoff bound coupled with the Bernstein's inequality \cite[Proposition 2.10,][]{wainwright2019high} to show that for any $\MMM\in\widetilde\CCC_{\MMM,\tttt}$, $\big| \langle \nabla l(\overline\MMM;\tttt),\MMM \rangle \big|$ still has a sub-Gaussian tail bound within the required scope, as is the case under the strong intensity condition.
\end{remark}

Furthermore, with a relatively large value of $R$, the optimization error $I_{3,\ddd}$ is dominated by $I_{1,\ddd}+I_{2,\ddd}$. Then, the convergence rate of $\|\MMM^{(R)}_{\ddd} - \MMM^*_{\ddd}\|_F^2$ is largely determined by the trade-off between $I_{1,\ddd}$ and $I_{2,\ddd}$. Corollary~\ref{cor:tensor1} specifies the convergence rate for the estimation error in the weak intensity regime. 

\begin{corollary}\label{cor:tensor1}
Suppose all the conditions in Theorem~\ref{thm:tensor1} are satisfied and $\log(nT)\prec T\prec\frac{n^2}{\log(nT)}$. 
Then, choosing $\frac{\sqrt{T}\log^{1/2}(nT)}{n}\prec\DDD_{\ddd}\prec1$, we have $$
\frac{1}{n_1n_2L}\|\MMM^{(R)}_{\ddd} - \MMM^*_{\ddd}\|_F^2 \preceq_P \frac{K_0}{L}. 
$$
\end{corollary}

\begin{remark}\label{rk:tensor1}
Corollary \ref{cor:tensor1} assures the validity of employing small intervals with $\DDD_{\ddd}\prec1$ in estimating the underlying tensor, to which the existing results \citep{han2022optimal, cai2023generalized} requiring the strong intensity assumption may not apply. It is also interesting to point out that the derived error bound in the weak and medium intensity regimes also provides practical guideline for network merging. By Corollary~\ref{cor:tensor1}, we will get a faster convergence rate as $\frac{K_0\log^{1/2+\epsilon}(nT)}{n\sqrt{T}}$ with $\DDD_{\ddd}\asymp\frac{\sqrt{T}\log^{1/2+\epsilon}(nT)}{n}$ or $L\asymp \frac{n\sqrt{T}}{\log^{1/2+\epsilon}(nT)}$, in contrast to the rate $\frac{K_0\log^{1+\epsilon}(nT)}{T}$ obtained in the strong intensity regime with $\DDD_{\ddd}\asymp\log^{1+\epsilon}(nT)$ or $L\asymp \frac{T}{\log^{1+\epsilon}(nT)}$ \citep{han2022optimal, cai2023generalized}.  The intuition is that if $T$ diverges very slowly, then one prefers to choose a relatively small $\DDD_{\ddd}$ or large $L$ to reduce the bias $I_{2,\ddd} = K_0\DDD_{\ddd}/T$. 
\end{remark}

\subsection{Error analysis based on adaptively merged intervals}\label{subsec:merge}

Define $\rho =\min_{k\in[{K_0}]}\|\ww^*_{k,\hhh}-\ww^*_{k-1,\hhh}\|$ and suppose $\rho\succeq 1$. Denote $r_{nT} = I_{1,\ddd}+I_{2,\ddd}$ as the upper bound in Theorem~\ref{thm:tensor1}. Theorem \ref{thm:cluster} shows that \eqref{eq:IC} gives a consistent estimate of ${K_0}$, and
\eqref{eq:cluster} further results in a precise recovery of the true partition $\hhh$ with overwhelming probability.

\begin{theorem}\label{thm:cluster} {\bf (Consistency of partition)}
Suppose all the conditions of Theorem~\ref{thm:tensor1} are satisfied, and $r_{nT} \prec \nu_{nT}\prec 1/{K_0}$. Then as $n$ and $T$ grow to infinity, we have $\Pr(\widehat{K} = {K_0}) \to 1$ and $\|\widehat\hhh-\hhh\|_{\infty} \preceq_P T r_{nT}$.
\end{theorem}

\begin{remark}\label{rk:cluster}
It is clear that the consistency of $\widehat{K}$ is guaranteed with a wide range of $\nu_{nT}$. Specifically, the condition $\nu_{nT}\prec 1/{K_0}$ implies that $\widehat{K} \geq {K_0}$, whereas $\nu_{nT}\succ r_{nT}$ guarantees $\widehat{K}\leq {K_0}$. More importantly, Theorem~\ref{thm:cluster} provides valuable guidelines for choosing $L$ and $\nu_{nT}.$ For fixed $K_0$, if $\log(nT)\prec T\prec \frac{n^2}{\log(nT)}$, we can choose $L = \frac{n\sqrt{T}}{\log^{1/2+\epsilon}(nT)}$ and $\nu_{nT} = \frac{\log^{1/4+\epsilon/2}(nT)}{n^{1/2}T^{1/4}}$; if $T\succeq\frac{n^2}{\log(nT)}$, we can choose $L = \frac{n\sqrt{T}}{\log^{3/2+\epsilon}(nT)}$ and $\nu_{nT} = \frac{\log^{3/4+\epsilon/2}(nT)}{n^{1/2}T^{1/4}}$. 
\end{remark}

Given that the true partition $\hhh$ is accurately estimated by $\widehat\hhh$, Theorem~\ref{thm:tensor2} further shows that the estimate $\MMM^{(R)}_{\widehat\hhh}$ based on the adaptively merged intervals $\widehat \hhh$ can attain a faster rate of convergence than that in Theorem \ref{thm:tensor1}.

\begin{theorem}\label{thm:tensor2} {\bf (Improved estimate via adaptive merging)}
Suppose all the conditions of Theorem~\ref{thm:cluster} are satisfied and $\DDD_{\hhh}\succeq \log^{2+\epsilon}(nK_0)$.
Then, with probability approaching 1, we have $$
\frac{1}{n_1n_2{K_0}}\|\MMM^{(R)}_{\widehat\hhh} - \MMM^*_{\hhh}\|_F^2   \preceq I_{1,\hhh}+I_{2,\hhh}+I_{3,\hhh},
$$ where $I_{1,\hhh} = \frac{1}{nT} + \frac{{K_0}}{n^2T}$, $$
I_{2,\hhh} = \left\{
\begin{aligned}
&K_0^2r_{nT}^2,&& \mbox{if }~Tr_{nT}\succ\log(nK_0),\\
&K_0^2r_{nT}^2\log^{2(1+\epsilon)}(nK_0),&& \mbox{if }~1\preceq Tr_{nT}\preceq\log(nK_0),\\
&\frac{K_0^2\log^{2(1+\epsilon)}(nK_0)}{T^2},&&\mbox{if }~Tr_{nT}\prec 1,
\end{aligned}
\right.
$$ 
and $I_{3,\hhh} = C(1 - \kappa)^R$ for some constants $C$ and $0<\kappa<1$.
\end{theorem}

Similarly, $I_{1,\hhh}$, $I_{2,\hhh}$ and $I_{3,\hhh}$ correspond to the estimation variance, the bias induced by adaptively merging, and the optimization error of \eqref{eq:update gene} after $R$ iterations, respectively. 
It is clear that $I_{1,\hhh}$ is much smaller than $I_{1,\ddd}$ in Theorem \ref{thm:tensor1} where the term $\frac{L}{n^2T}$ is reduced to $\frac{K_0}{n^2T}$. The convergence rate for the bias term, $I_{2,\hhh}$, takes different forms depending on the term $Tr_{nT}$. Specifically,  Corollary~\ref{cor:tensor2} gives the convergence rate for the estimation error of $\MMM^{(R)}_{\widehat\hhh}$.

\begin{corollary}\label{cor:tensor2}
Suppose all the conditions in Theorem~\ref{thm:tensor2} are satisfied. If $T\succeq\frac{n^2}{\log(nT)}$, then choosing $\DDD_{\ddd}\succ\log(nT)$ leads to 
$$
\frac{1}{n_1n_2K_0}\|\MMM^{(R)}_{\widehat\hhh} - \MMM^*_{\hhh}\|_F^2 \preceq_P \frac{1}{nT}+\frac{K_0}{n^2T}+\frac{K_0^2L^2}{n^4T^2}+\frac{K_0^4}{L^2};
$$ 
if $\log(nT)\prec T\prec\frac{n^2}{\log(nT)}$, choosing $\frac{\sqrt{T}\log^{1/2}(nT)}{n}\prec \DDD_{\ddd}\prec 1$ leads to  
$$
\frac{1}{n_1n_2K_0}\|\MMM^{(R)}_{\widehat\hhh} - \MMM^*_{\hhh}\|_F^2 \preceq_P \frac{1}{nT}+\frac{K_0}{n^2T}+\frac{K_0^2\log^{2(1+\epsilon)}(nK_0)}{T^2}.
$$
\end{corollary}

\begin{remark}\label{rk:tensor2}
Let $K_0$ be a fixed constant, and we compare the estimates $\MMM^{(R)}_{\ddd}$ based on the equally spaced intervals and $\MMM^{(R)}_{\widehat\hhh}$ based on the adaptively merged intervals. If $T\succeq\frac{n^2}{\log(nT)}$, then $\MMM^{(R)}_{\widehat\hhh}$ converges to 0 at a faster rate of $\frac{1}{nT}+\frac{\log^{3+2\epsilon}(nT)}{n^2T}$, whereas the convergence rate of $\MMM^{(R)}_{\ddd} $ with $L = \frac{n\sqrt{T}}{\log^{3/2+\epsilon}(nT)}$ is of order $\frac{\log^{3/2+\epsilon}(nT)}{n\sqrt{T}}$. If $\log(nT)\prec T\prec \frac{n^2}{\log(nT)}$, the convergence rates of $\MMM^{(R)}_{\widehat\hhh}$  and $\MMM^{(R)}_{\ddd}$ are of order $\frac{1}{nT}+\frac{\log^{2(1+\epsilon)} n}{T^2}$ and $\frac{\log^{1/2+\epsilon}(nT)}{n\sqrt{T}}$ with $L = \frac{n\sqrt{T}}{\log^{1/2+\epsilon}(nT)}$, 
where $\MMM^{(R)}_{\widehat\hhh}$ is still advantageous as long as $T\succ n^{\frac{2}{3}}\log^{1+\frac{2}{3}\epsilon}(nT)$.
\end{remark}

Table~\ref{tab:res} summarizes the convergence rates of the proposed method. It is shown that in all scenarios of $n$ and $T$, if we suppress the logarithm terms, the optimal $L$ is always of order $n\sqrt{T}$.  When $T\succeq\frac{n^2}{\log(nT)}$, the optimal choice of $\DDD_{\ddd}\succ\log(nT)$ makes the initial estimate fall into the strong intensity regime. When $T\prec\frac{n^2}{\log(nT)}$, the optimal choice of $L$ makes the initial estimate fall into the weak intensity regime, and adaptive merging will further improve the convergence rate as long as $T\succ n^{\frac{2}{3}}\log^{1+\frac{2}{3}\epsilon}(nT)$. Though this advantage will vanish when $T\preceq n^{\frac{2}{3}}\log^{1+\frac{2}{3}\epsilon}(nT)$, in which case the initial estimate $\widehat\MMM_{\ddd}$ would be a better choice. Note that the error rates for the proposed method are always smaller than the rates obtained in the strong intensity regime based on equally spaced intervals, which are $\frac{\log^{3/2+\epsilon}(nT)}{n\sqrt{T}}$ in the first scenario and $\frac{\log^{1+\epsilon}(nT)}{T}$ in the second and third scenarios.


\begin{table}[!h]
\begin{center}
\begin{footnotesize}
\begin{tabular}{c|c|c|c|c } 
\hline
\hline
 Scenarios for $(n,T)$ & Optimal $L$ & Intensity & Merging & Error Rates   \\
\hline
\hline
  $T\succeq\frac{n^2}{\log(nT)}$ & $\frac{n\sqrt{T}}{\log^{3/2+\epsilon}(nT)}$ & Strong &Yes& $\frac{1}{nT}+\frac{\log^{3+2\epsilon} (nT)}{n^2T}$  \\ \cline{1-5} 
 $n^{\frac{2}{3}}\log^{1+\frac{2}{3}\epsilon}(nT)\prec T\prec \frac{n^2}{\log(nT)}$  & $\frac{n\sqrt{T}}{\log^{1/2+\epsilon}(nT)}$ &Weak & Yes & $\frac{1}{nT}+\frac{\log^{2(1+\epsilon)}(nT)}{T^2}$  \\ \cline{1-5} 
 $\log(nT)\prec T\preceq n^{\frac{2}{3}}\log^{1+\frac{2}{3}\epsilon}(nT)$ & $\frac{n\sqrt{T}}{\log^{1/2+\epsilon}(nT)}$ &Weak & No & $\frac{\log^{1/2+\epsilon}(nT)}{n\sqrt{T}}$    \\ \cline{1-5} 
\hline
\hline
\end{tabular}
\end{footnotesize}
\end{center}
\caption{Convergence rates for the proposed method in different regimes.}
\label{tab:res}
\end{table}

\section{Numerical experiments}

\subsection{Simulation examples}
\label{sec:simu}

We let $n_1 = n_2 = n \in\{50, 100\}$ and $T \in\{ n^2 / \log n, n, n^{1/3}\}$, corresponding to three scenarios in Table~\ref{tab:res}. We set $r_1 = r_2 = r_3 = 3$, $K_0 \in\{3,5\}$ and the partition $\hhh\in\R^{K_0}$ is constructed in such a way that each $\eta_k$ is randomly generated from $[0,T)$, where the length ratio for the largest and smallest intervals is no larger than 3, and $\ww^*(t)$ is a piecewise constant function of $t$ with $\ww^*(t) = (\WW_{\hhh}^*)_{[k, ]}$ for $t\in[\eta_{k-1},\eta_{k})$. The columns of $\WW_{\hhh}^*$ are randomly generated such that $\int_{0}^T \ww^*(t)\ww^*(t)^\top dt = T\II_{2}$, while the columns of $\UU^*/\sqrt{n_1}$ and $\VV^*/\sqrt{n_2}$ are generated uniformly from $\OOO_{n,2}$. For $\SSSS^*$, the diagonal entries are set to be 0.5 and the rest entries 0.

We investigate the finite-sample performance of the proposed method, and compare it with existing tensor decomposition methods, including a modified Poisson tensor PCA \citep{han2022optimal}, higher-order orthogonal iteration \citep{de2000best} and higher-order SVD \citep{de2000multilinear}.
Specifically, we denote 
\begin{itemize}
\item AM($\widehat K$) as the estimate based on adaptively merged intervals;
\item ES($L_{\opt}$) as the proposed initial estimate built on $L_{\opt}$ equally spaced intervals, where $L_{\opt}$ is based on Table~\ref{tab:res};
\item ES($L_{\str}$) as the estimate based on $L_{\str}\asymp \frac{T}{\log^{1+\epsilon}(nT)}$ equally spaced intervals in the strong intensity regime; 
\item ``HOOI'' and ``HOSVD'' as the estimates of higher-order orthogonal iteration \citep{de2000best} and SVD \citep{de2000multilinear} based on $L_{\str}$ equally spaced intervals, where $(\YYY_{\ddd})_{ijl} = |\TT_{ij}\cap[\delta_{l-1},\delta_l)|$. 
\end{itemize}
Their numeric performance is assessed by the average tensor estimation error based on the corresponding intervals. 

The averaged tensor estimation errors over 50 independent replications and their standard errors for each method are summarized in Tables \ref{tab:error K3} and \ref{tab:error K5}. It is shown that AM($\widehat K$) has delivered superior numerical performance and outperforms the other three competitors in the first two scenarios, $T = n^2 / \log n$ and $T = n$, in all examples, which is consistent with the theoretical results in Table~\ref{tab:res}. It is interesting to note that in the third scenarios where $T = n^{1/3}$, ES($L_{\opt}$) outperforms AM($\widehat K$), which echos the results in Theorem~\ref{cor:tensor2} and Reamrk~\ref{rk:tensor2}. 
It is worthy pointing out that AM($\widehat K$) and ES($L_{\opt}$) show great advantage over ES($L_{\str}$) and HOOI in all scenarios, suggesting superiority of the proposed method. Further, Tables \ref{tab:error K3} and \ref{tab:error K5} also show that $K$ could be consistently selected by \eqref{eq:IC}. 

\begin{figure}
    \centering
    \begin{minipage}[b]{0.32\linewidth}
        \centering
        \includegraphics[width=\textwidth]{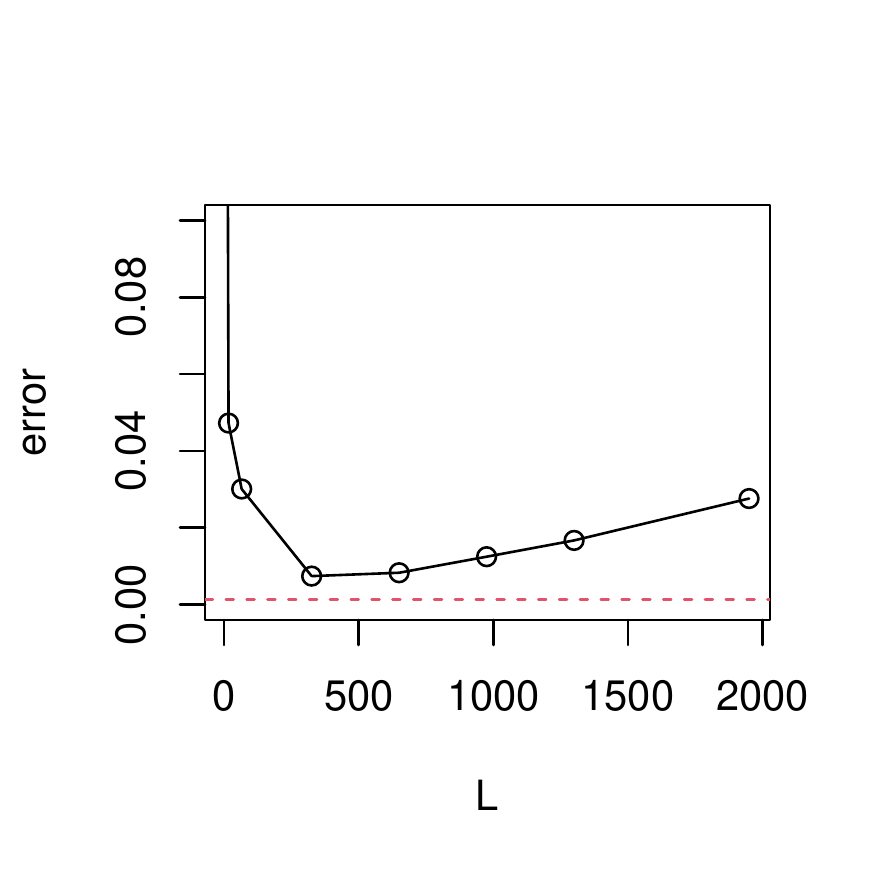}
    \end{minipage}
    \begin{minipage}[b]{0.32\linewidth}
        \centering
        \includegraphics[width=\textwidth]{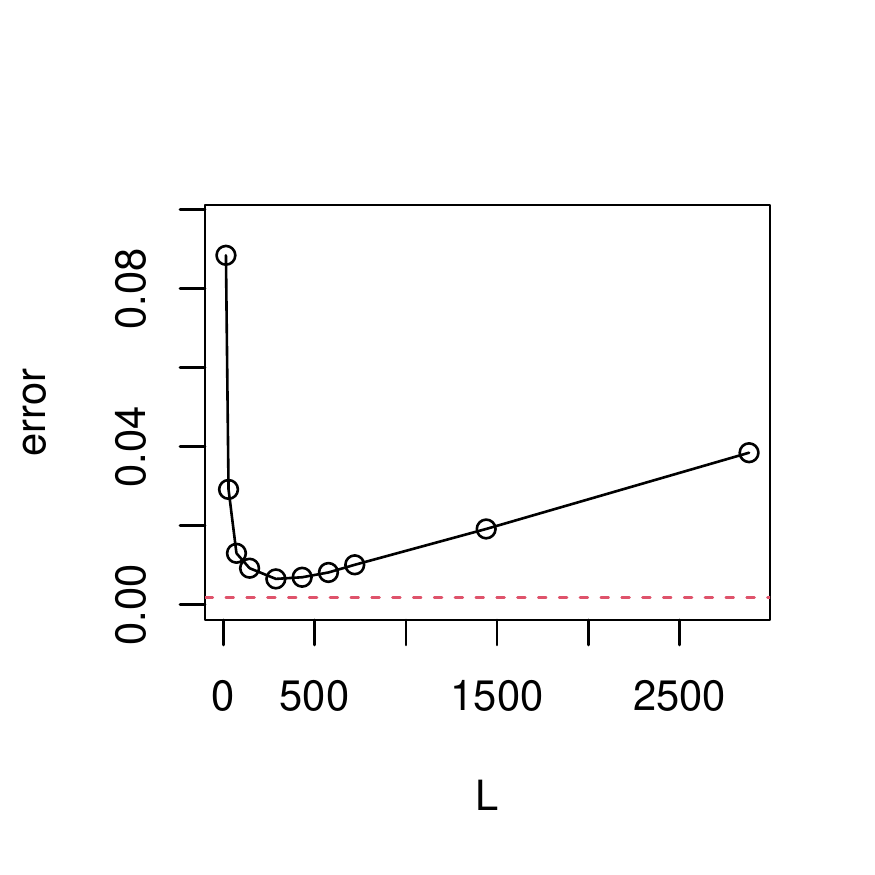}
    \end{minipage}
    \begin{minipage}[b]{0.32\linewidth}
        \centering
        \includegraphics[width=\textwidth]{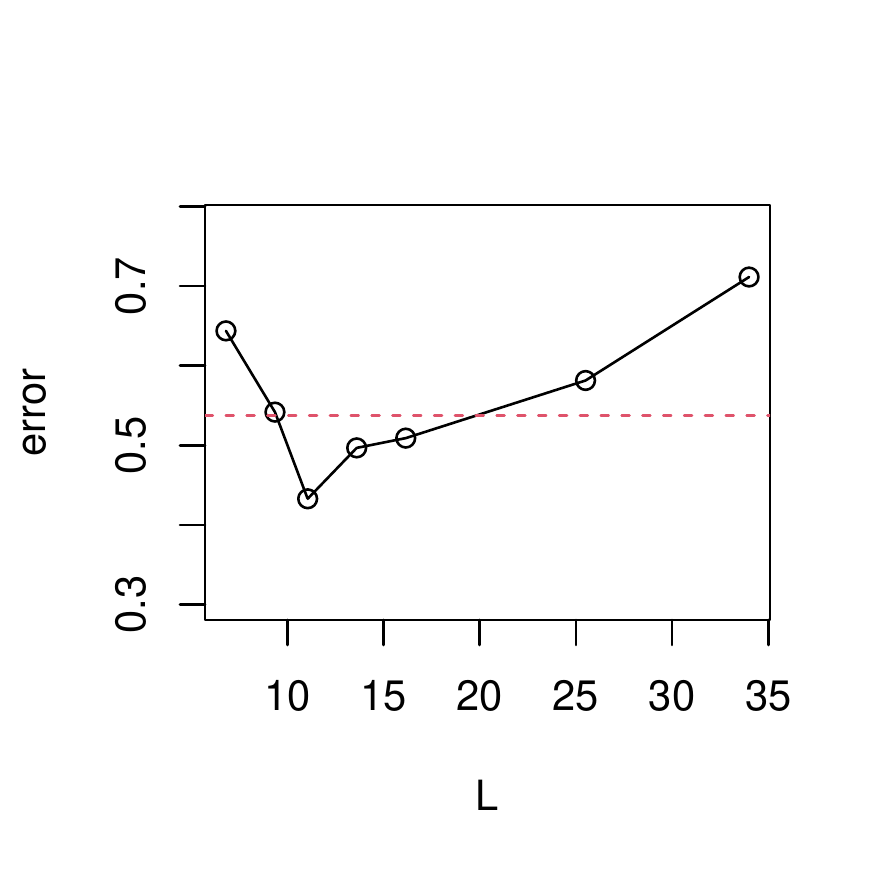}
    \end{minipage}
	\caption{The average tensor estimation errors based on equal spaced intervals under three scenarios of Table~\ref{tab:res} with different values of $L$ over 50 independent replications. 
	The red dotted lines are the average estimation errors of the estimate based on adaptively merged intervals. The large error rates in the third panel is due to the much smaller chosen $T$ in the third scenario.}
\label{fig:compare}
\end{figure}

We now scrutinize how the tensor estimation error is affected by different choices of $L$ in examples with $n = 50$, $T \in\{ n^2 / \log n, n, n^{1/3}\}$ and $K_0 = 3$.
The three panels of Figure~\ref{fig:compare} show the average tensor estimation errors of ES($L$) over 50 independent replications with different $L$. Clearly, as $L$ increases, the error decreases at first, and then increases. This is because the bias induced by the partition with a small number of intervals dominates the tensor estimation error in each interval, which will be reduced dramatically as $L$ increases. Yet, as $L$ becomes larger, the estimation variance begins to dominate the tensor estimation error, and it increases along with $L$. This phenomenon validates the asymptotic upper bound in Theorem~\ref{thm:tensor1}. The averaged tensor estimation error of AM$(\widehat K)$ with $\widehat K= 3$ adaptively merged intervals is represented by the red dotted line, which is smaller than that of all the methods based on equally spaced intervals in the first two scenarios, demonstrating the advantage of the proposed methods in Theorem~\ref{thm:tensor2}. However, in the third scenario, AM$(\widehat K)$ is defeated by ES($L_{\opt}$) for certain $L$, which also validates the results in Theorem~\ref{cor:tensor2} and Reamrk~\ref{rk:tensor2}. It suggests that the initial estimate $\widehat\MMM_{\ddd}$ would be a better choice when $T\preceq n^{\frac{2}{3}}\log^{1+\frac{2}{3}\epsilon}(nT)$ as shown in the Table~\ref{tab:res}.


\begin{table}[!htb]
	\begin{center}
		\caption{The averaged tensor estimation errors and $\widehat K$, when $K_0=3$}
		\label{tab:error K3}
		\begin{small}
			\begin{tabular}{ c|c|c|c|c } 
				\hline
				\hline
				Error& Method & $T = n^2 / \log n$~~ & $T = n$~~ & $T = n^{1/3}$~~ \\
				\hline
				\hline
				\multirow{6}{4em}{$n = 50$} & $\widehat K$  &3(0) & 3(0) & 3(0) \\ \cline{2-5} 
				& AM($\widehat K$) &0.0014(0.0001) & 0.0017(0.0002) & 0.5316(0.2) \\ \cline{2-5} 
				& ES($L_{\opt}$) &0.0075(0.0003) & 0.0065(0.0003) & 0.4318(0.2) \\ \cline{2-5} 
				& ES($L_{\str}$) &0.0075(0.0003) & 0.268(0.002) & 1.0241(0.7) \\ \cline{2-5} 
				& HOOI  &1.9976(0.001) & 0.2215(0.003) & 6.6766(0.01) \\ \cline{2-5} 
				& HOSVD  &2.0405(0.004) & 0.2206(0.003) & 6.6963(0.01) \\ \cline{2-5} 
				\hline
				\hline
				\multirow{6}{4em}{$n=100$} & $\widehat K$  &3(0) & 3(0) & 3(0) \\ \cline{2-5} 
				& AM($\widehat K$) &0.0002(2e-05) & 0.0004(3e-05) & 0.1371(0.03) \\ \cline{2-5} 
				& ES($L_{\opt}$) &0.0006(5e-05) & 0.0015(7e-05) & 0.1175(0.007) \\ \cline{2-5} 
				& ES($L_{\str}$) &0.0006(5e-05) & 0.1325(0.0006) & 0.3032(0.01) \\ \cline{2-5} 
				& HOOI &1.6588(0.0003) & 0.1064(0.0005) & 5.6621(0.005)\\ \cline{2-5} 
				& HOSVD &1.6656(0.0007) & 0.1066(0.0005) & 5.6641(0.004) \\ \cline{2-5} 
				\hline
				\hline
			\end{tabular}
		\end{small}
	\end{center}
\end{table}

\begin{table}[!htb]
	\begin{center}
		\caption{The averaged tensor estimation errors and $\widehat K$ when $K_0=5$}
		\label{tab:error K5}
		\begin{small}
			\begin{tabular}{ c|c|c|c|c } 
				\hline
				\hline
				Error& Method & $T = n^2 / \log n$~~ & $T = n$~~ & $T = n^{1/3}$~~ \\
				\hline
				\hline
				\multirow{6}{4em}{$n = 50$} & $\widehat K$  &5(0) & 5(0) & 4.5(0.7) \\ \cline{2-5} 
				& AM($\widehat K$) &0.0013(0.0001) & 0.0016(0.0002) & 0.7052(0.4) \\ \cline{2-5} 
				& ES($L_{\opt}$) &0.0069(0.0003) & 0.0075(0.001) & 0.3551(0.04)\\ \cline{2-5} 
				& ES($L_{\str}$) &0.0069(0.0003) & 0.0502(0.0009) & 1.2808(1) \\ \cline{2-5} 
				& HOOI &1.9662(0.001) & 0.4487(0.002) & 9.3893(0.007) \\ \cline{2-5} 
				& HOSVD &2.0214(0.002) & 0.4213(0.006) & 9.3983(0.007) \\ \cline{2-5} 
				\hline
				\hline
				\multirow{6}{4em}{$n=100$} & $\widehat K$  &5(0) & 5(0) & 4.8(0.5)\\ \cline{2-5} 
				& AM($\widehat K$) &0.0002(2e-05) & 0.0008(4e-05) & 0.2997(0.3)\\ \cline{2-5} 
				& ES($L_{\opt}$) &0.0039(6e-05) & 0.0031(8e-05) & 0.0592(0.02) \\ \cline{2-5} 
				& ES($L_{\str}$) &0.0039(6e-05) & 0.1721(0.001) & 0.3666(0.03)\\ \cline{2-5} 
				& HOOI  &3.2263(0.0003) & 0.127(0.0007) & 9.5284(0.004)\\ \cline{2-5} 
				& HOSVD  &3.2381(0.0005) & 0.1309(0.0007) & 9.53(0.004)\\ \cline{2-5} 
				\hline
				\hline
			\end{tabular}
		\end{small}
	\end{center}
\end{table}

\subsection{Real example}

We apply the proposed method to analyze a longitudinal network based on the militarized interstate dispute dataset \citep{palmer2022mid5}. The dataset consists of all the major interstate disputes and  involved countries during 1895-2014. It can be converted into a longitudinal network with nodes representing all countries ever involved in any dispute over the years. Particularly, we set $dy_{ij}(t) = 1$ if country $i$ cooperated with country $j$ in a militarized interstate dispute occurred at time $t$. We keep it as $1$ for the following years until a dispute occurred between themselves, and then $dy_{ij}(t)$ changes to 0 and remains until the next cooperation. This pre-processing step leads to a longitudinal network with $n_1=n_2=195$ nodes and $110066$ temporal edges, and the time stamps range from 0 to $T = 120$ years. We apply the proposed method with $\DDD_{\ddd} = 5$ years and thus $L = 24$, where the ranks are set to be $(r_1,r_2,r_3) = (2,2,2)$ following a similar rank selection procedure in \cite{han2022optimal}.

To assess the numeric performance, we randomly split the node pairs into 5 disjoint subsets $\{\PPP_p\}_{p=1}^5$. For each $p$, we obtain the estimated tensor $\widehat\MMM^{(p)}$ on $\PPP_{-p} = [n_1]\times[n_2]\backslash\PPP_p$, and validate the estimation accuracy on $\PPP_p$ in each small interval by,
$$
\text{err}^{(p)} = \frac{ \|(\TT-\widehat\YYY^{(p)}) \circ {\boldsymbol1}_{\PPP_p}\|_F }{ \|\TT \circ {\boldsymbol1}_{\PPP_p}\|_F},
$$ where $\TT = (\TT_{ij})_{n_1\times n_2}$ and $\widehat\YYY^{(p)}\in\R^{n_1\times n_2}$ contain the true and estimated numbers of temporal edges for each node pair $(i,j)$, ${\boldsymbol1}_{\PPP_p}\in\R^{n_1\times n_2}$ is the indicator matrix for $\PPP_p$, and $\circ$ denotes the matrix Hadamard product. Then, the testing error is calculated as $\text{err} = \sum_{p=1}^5\text{err}^{(p)}/5$. For AM($\widehat K$), $\widehat\YYY^{(p)}_{\widehat\hhh}$ is obtained by $(\widehat\YYY_{\widehat\hhh}^{(p)})_{ij} =\sum_{k=1}^{\widehat K} \lambda_0\exp((\widehat\MMM_{\widehat\hhh}^{(p)})_{ijk})(\widehat\eta_k-\widehat\eta_{k-1})$, whereas
$(\widehat\YYY_{\ddd}^{(p)})_{ij} =\sum_{l=1}^{\widehat L} \lambda_0\exp((\widehat\MMM_{\ddd}^{(p)})_{ijk})(\delta_l-\delta_{l-1})$ for ES($L_{\opt}$). The estimates by HOSVD and HOOI are obtained in the same way as in Section~\ref{sec:simu}. The averaged testing errors and their standard errors for the competing methods over 50 times replications are provided in Table~\ref{tab:war}. It is evident that AM($\widehat K$) and ES($L$) significantly outperform the spectral methods, and the difference between AM($\widehat K$) and ES($L$) is not significant, which is not surprising as $T = 120$ is not large enough compared with $n=195$, corresponding to the third scenario in Table \ref{tab:res}. 

\begin{table}[!h]
\begin{center}
\caption{The average testing errors and standard errors (in parentheses) for various methods over 50 replications.}
\label{tab:war}
\begin{small}
\begin{tabular}{c|c|c|c}
\hline
\hline
AM($\widehat K$) & ES(L) & HOSVD & HOOI \\
\hline
0.739(0.037) & 0.752(0.087) & 1.160(0.002) & 1.163(0.002)\\
  \hline
\hline
\end{tabular}
\end{small}
\end{center}
\end{table}

\begin{figure}[!htb]
	\centering
        \includegraphics[width=0.9\textwidth]{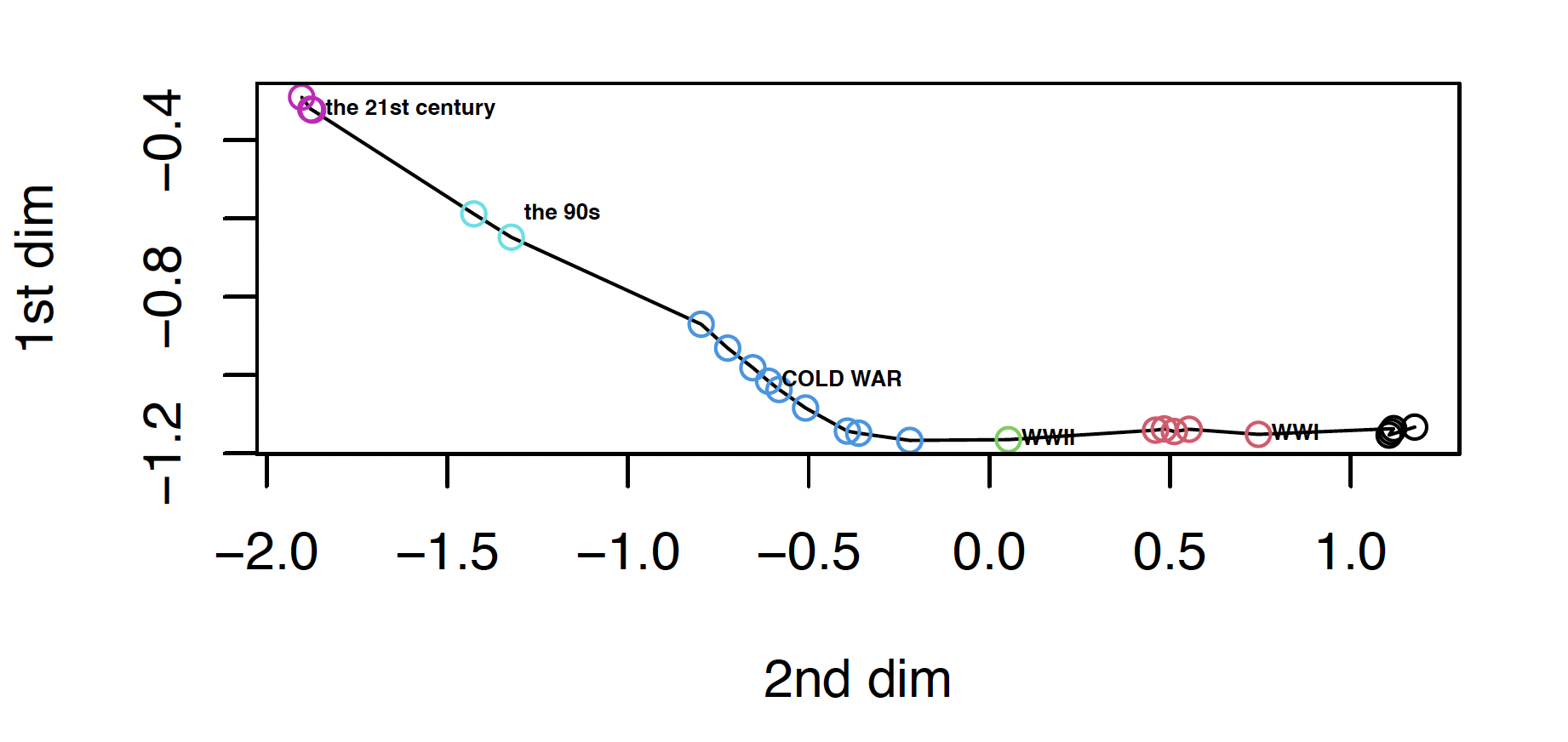}
	 \caption{The estimated temporal embedding vectors $\{\widehat\ww_{l,\ddd}\}_{l=1}^9$, where colors represent different merged time intervals.}
 \label{fig:results}
\end{figure}

Furthermore, the output of AM$(\widehat K)$ yields that $\widehat K = 6$ and $\widehat\hhh = (20, 45, 50, 95, 105, 120)$, and thus the adaptively merged time intervals are 1895-1914, 1915-1939, 1940-1944, 1945-1989, 1990-1999 and 2000-2014. These intervals appear to be closely related with a number of major world-wide events: before WWI, recess between WWI and WWII, WWII, Cold War, the 90s, and the 21st century. The estimated temporal embedding vectors $\{\widehat\ww_{l,\ddd}\}_{l=1}^L$ are shown in Figure~\ref{fig:results}, where $\widehat\ww_{l,\ddd}$ in different merged time intervals, represented by different colors, are well separated.


\begin{figure}[!htb]
	\centering
         \includegraphics[width=0.9\textwidth]{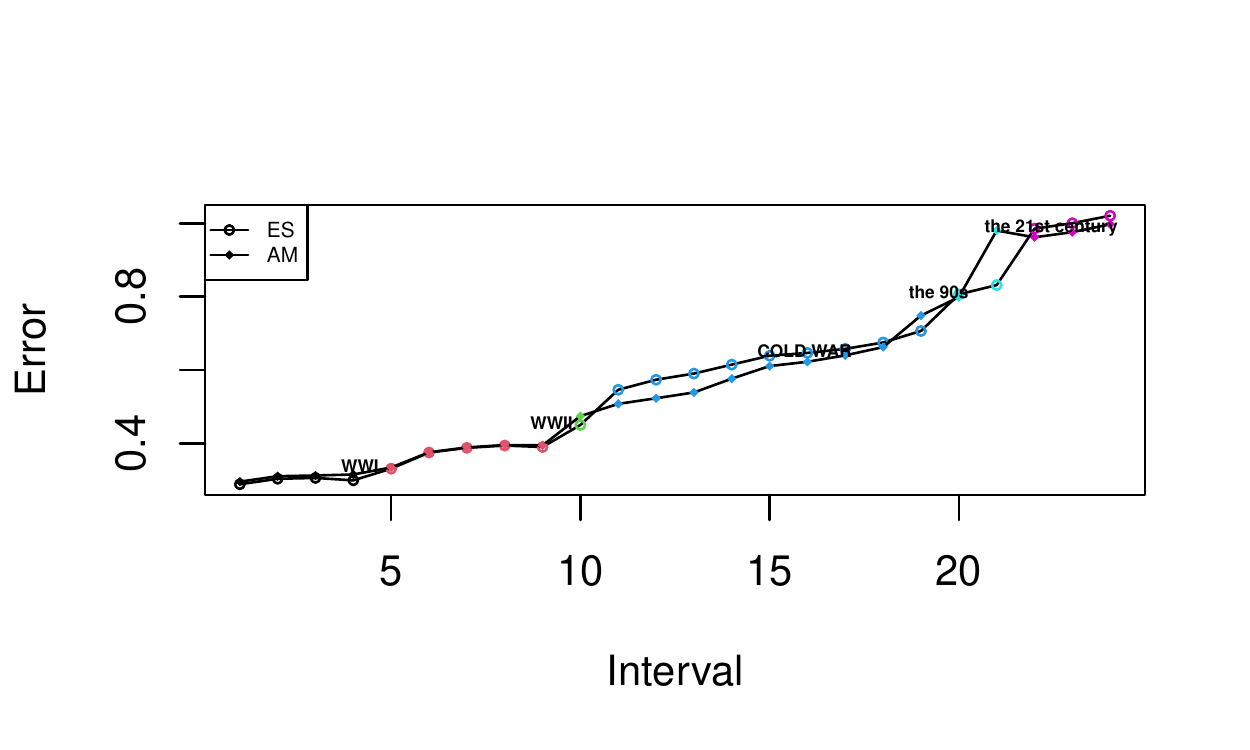}
	 \caption{The average estimation errors of AM($\widehat K$) and ES($L$) in each of the $L=24$ intervals over 50 replications. 
}
 \label{fig:error war}
\end{figure}

It is also interesting to examine the averaged estimation error in each small intervals $[\delta_{l-1},\delta_l)$, as displayed in Figure~\ref{fig:error war}. Clearly, the estimation errors of AM($\widehat K$) are generally smaller than ES($L$) in intervals that do not contain the estimated change points, but more or less comparable in intervals containing the estimated change points. This phenomenon reveals that adaptive merging actually leads to a smaller tensor estimation error than ES($L$) over the time line, while it produces similar errors in those small number of intervals containing the estimated change points, which somehow dominates the tensor estimation errors.

\section{Discussion}

In this paper, we propose an efficient estimation framework for longitudinal network, leveraging strengths of adaptive network merging, tensor decomposition and point process. A thorough analysis is conducted to quantify the asymptotic behavior of the proposed method, which shows that adaptively network merging leads to substantially improved estimation accuracy compared with existing competitors in literature. The theoretical analysis also provides a guideline for network merging under various scenarios. The advantage of the proposed method is supported in the numerical experiments on both synthetic and real longitudinal networks. The proposed estimation framework can be further extended to incorporate edge-wise or node-wise covariates or employ some more general counting processes, which will be left for future investigation.

\section*{Acknowledgment}
This research is supported in part by HK RGC Grants GRF-11304520, GRF-11301521, GRF-11311022, and CUHK Startup Grant 4937091.

\bibliographystyle{apalike}
\bibliography{ref}

\end{document}